\documentclass[journal]{IEEEtran}
\usepackage{ctable}
\usepackage{color}
\usepackage{multirow}
\usepackage{enumitem}
\usepackage{amsfonts}
\usepackage[font=small,labelfont=bf,justification=justified]{caption}

\definecolor{myred}{RGB}{255, 0, 0}
\definecolor{mygreen}{RGB}{0, 165, 60}
\definecolor{myblue}{RGB}{0, 112, 192}

\usepackage{cite}
\usepackage{amsmath}
\ifCLASSOPTIONcompsoc
  \usepackage[caption=false,font=normalsize,labelfont=sf,textfont=sf]{subfig}
\else
  \usepackage[caption=false,font=footnotesize]{subfig}
\fi
\usepackage{url}

\begin{document}
\bstctlcite{IEEEexample:BSTcontrol}
%
\title{Exploring Cross-Domain Pretrained Model\\for Hyperspectral Image Classification}
%
%
%

\author{Hyungtae Lee,~\IEEEmembership{Member,~IEEE,}
        Sungmin Eum,~\IEEEmembership{Member,~IEEE,}
        and~Heesung Kwon,~\IEEEmembership{Senior Member,~IEEE}
\thanks{Manuscript received April 19, 2005; revised August 26, 2015.}
\thanks{H. Lee and H. Kwon are with the Intelligent perception branch, the Computational \& Information Sciences Directorate (CISD), Army Research Laboratory, Adelphi, MD, 20783 USA (e-mail: \{hyungtae.lee,~heesung.kwon\}.civ@mail.mil).}
\thanks{S. Eum is with Booz Allen Hamilton Inc., McLean, VA, 22102 USA and with the Intelligent perception branch, the Computational \& Information Sciences Directorate (CISD), Army Research Laboratory, Adelphi, MD, 20783 USA (e-mail: eum\_sungmin@bah.com).}
\thanks{© 2022 IEEE. Personal use of this material is permitted. Permission from IEEE must be obtained for all other uses, in any current or future media, including reprinting/republishing this material for advertising or promotional purposes, creating new collective works, for resale or redistribution to servers or lists, or reuse of any copyrighted component of this work in other works.}}

%
%

\markboth{IEEE Transactions on Geoscience and Remote Sensing Submission}%
{Lee \MakeLowercase{\textit{et al.}}: Building Pre-trained Model for Hyperspectral Image ClassificationExploring Cross-Domain Pretrained Model for Hyperspectral Image Classification}
%



\maketitle

\begin{abstract}
A \emph{pretrain-finetune strategy} is widely used to reduce the overfitting that can occur when data is insufficient for CNN training. First few layers of a CNN pretrained on a large-scale RGB dataset are capable of acquiring general image characteristics which are remarkably effective in tasks targeted for different RGB datasets. However, when it comes down to hyperspectral domain where each domain has its unique spectral properties, the pretrain-finetune strategy no longer can be deployed in a conventional way while presenting three major issues: 1) inconsistent spectral characteristics among the domains (e.g., frequency range), 2) inconsistent number of data channels among the domains, and 3) absence of large-scale hyperspectral dataset.

We seek to train a universal \emph{cross-domain} model which can later be deployed for various spectral domains. To achieve, we physically furnish multiple inlets to the model while having a universal portion which is designed to handle the inconsistent spectral characteristics among different domains. Note that only the universal portion is used in the finetune process. This approach naturally enables the learning of our model on multiple domains simultaneously which acts as an effective workaround for the issue of the absence of large-scale dataset.

We have carried out a study to extensively compare models that were trained using \emph{cross-domain} approach with ones trained from scratch. Our approach was found to be superior both in accuracy and in training efficiency. In addition, we have verified that our approach effectively reduces the overfitting issue, enabling us to deepen the model up to 13 layers (from 9) without compromising the accuracy.
\end{abstract}


%
\IEEEpeerreviewmaketitle

\section{Introduction}
%
%
%
%

\IEEEPARstart{I}{n} many classification tasks, convolutional neural network (CNN) has been showing a series of innovative performances. However, when only given a small-sized target dataset, it is difficult to avoid the overfitting issue due to an enormous number of parameters that need to be optimized in a deep CNN. One widely known approach to go around this issue is to finetune the model from first few layers of a pretrained model which was previously trained on a large-scale dataset~\cite{RGirshickTPAMI2016}. This approach can be applied effectively when the source and the target datasets share equivalent spectral characteristics (e.g., RGB to RGB) and the size of the source dataset is much larger than that of the target dataset.

\begin{figure}[t]
\centering
\includegraphics[width=0.95\linewidth,trim=30mm 68mm 35mm 68mm,clip]{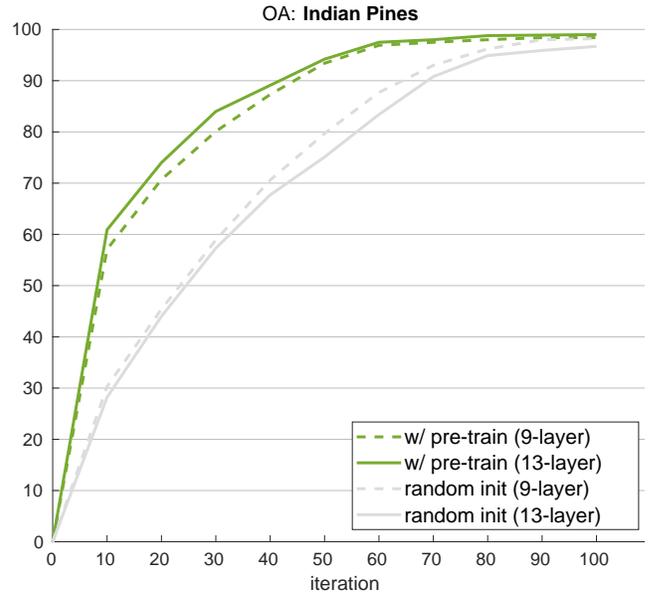}
\caption{{\bf Learning curve on OA.} 9- and 13-layer models initialized from scratch or a pretrained model are evaluated with Indian Pines domain. The pretrained model is trained on six source domains by using our cross-domain approach. Evaluation metric is overall accuracy (OA).}
\label{fig:ap_evolution_intro}
\end{figure}

However, when considering classification tasks in hyperspectral domain\footnote{In this paper, ``hyperspectral domain'' is represented by a dataset consisting of hyperspectral data (typically one single image) and its associated classification task.} where domains contain their own spectral properties, there are challenges when using this conventional {\it pretrain-finetune} strategy. First of all, hyperspectral datasets acquired by different sensors have different spectral characteristics. For example, Indian Pines domain has 200 bands representing 0.4$\sim$2.5$\mu$m frequency range while Pavia Centre domain has 102 bands covering 0.43$\sim$0.86$\mu$m. The spectral feature learned from the source may be unfit for the target usage. Moreover, as number of data channels among different hyperspectral domains are inconsistent, it is infeasible to build an initial layer which can universally handle different domains. Lastly, due to difficulties in annotating hyperspectral images, most of the domains only contain an extremely small number of labeled examples ($\sim$10K), which can easily cause overfitting issues when training high-capacity CNN models.

To address the aforementioned issues, we devise a \emph{cross-domain} approach which enables the training of a CNN model which can be universally deployed for various spectral domains. We have equipped this model with multiple inlets (i.e., data analysis layers) to physically handle different domains. Meanwhile, in order to capture the universal aspect of different domains, we prepare a shared portion in the model (i.e., mid-level feature analysis layer). Task-specific layers can be appended on the other side of the shared portion of the model dependent upon the task properties such as category size. As the overall design of the architecture allows simultaneous training of all the available hyperspectral domains, the issue regarding the absence of large-scale dataset can be remedied effectively. 

In the overall finetune process for a given target domain, only the shared portion of the model pretrained on multiple source domains is to be inherited while all the other layers (data analysis and task-specific layers) are initialized and trained from scratch. It is noteworthy to mention that using a 10$\times$ learning rate for the data analysis layer (while keeping 1$\times$ for the shared portion) was crucial to obtain a significant accuracy gain. This specific procedure can quickly adapt the data analysis layer to the newly introduced target domain while slowly optimizing the shared portion of the model which already contained the general spectral characteristics acquired from the source domain.

In this paper, we validate the effectiveness of using a cross-domain pretrain-finetune approach for the task of hyperspectral image classification. We have first designed a CNN model which outperforms all the available CNN-based models by employing recently introduced performance-increasing modules. Across all the experiments, this model has been set as the backbone model to evaluate the performance with/without the cross-domain pretrain-finetune approach. Based on our experiments, we observed several benefits of using a cross-domain pretrained model, as shown in Figure~\ref{fig:ap_evolution_intro}. Most importantly, the pretraining provides better accuracy than its counterpart randomly initialized from scratch. In addition, the proposed approach moderates the overfitting issue which can occur when the network depth increases, allowing deeper layers (up to 13 layers from 9) without compromising accuracy. We also observed that our approach results in faster training convergence which can reduce the training time.

To provide a practical set of guides in training an effective cross-domain pretrained CNN model, we carried out an additional comprehensive study to answer the following questions\footnotemark: 
\begin{enumerate}
    \item \textit{Is pretraining necessary for hyperspectral image classification?}
    \item \textit{Does having a larger source domain improve overall accuracy?}
    \item \textit{Is pretraining effective when target and source domains are obtained from different sensors?}
    \item \textit{Does introducing more variety in the source domains for pretraining increase the accuracy?}
\end{enumerate}
\footnotetext{Although we elaborated such practical research questions within our preliminary work~\cite{HLeeIGARSS2019}, we managed to analyze our pretrain-finetune approach with a limited number of experimental evaluations and thus could not address the resolutions for the questions in a thorough manner. In this journal manuscript, the entire set of experiments has been redesigned to reach at the conclusions through more comprehensive analyses.

While the initial form of the cross-domain architecture used for the pretraining process was introduced in another preliminary work ~\cite{HLeeIGARSS2018}, a restructuring procedure was carried out in order to suit the need for our final pretrain-finetune process. In addition, we constructed a completely different backbone yielding a higher accuracy. We have also conducted an extensive set of ablation studies, and newly added experimental comparisons with additional set of CNN-based hyperspectral image classification models. Instead of appending additional experiments and analyses on the aforementioned manuscripts, substantial portion of this manuscript has been rewritten.}

\section{Related Works}
\label{sec:related_work}

\subsection{Pretrain and Finetune}
\label{ssec:pretraining}

Girshick et al.~\cite{RGirshickTPAMI2016} firstly use a pretrained model trained on very large-scale dataset to overcome data scarcity. At that time, existing object detection dataset (e.g., PASCAL VOC~\cite{MEveringhamIJCV2010}) did not include enough images annotated with object information (bounding boxes, categories, etc.). Therefore, the state-of-the-art CNN architecture for image classification becomes the backbone of object detection model, and the model weights are finetuned from the backbone trained on a very large-scale ImageNet dataset~\cite{ORussakovskyIJCV2015}. In~\cite{RGirshickTPAMI2016}, this strategy increases the accuracy by 8.0\% on PASCAL VOC. Due to such a large margin of accuracy, pretrained models have been used undoubtedly in many tasks~\cite{ABencyECCV2016,KHeCVPR2020,HLeeTIP2020,HLeeICASSP2018,HLeeArXiv2016,MOquabCVPR2014,CRealeCVPRW2017} over the past few years.

Recently several researchers have begun comprehensive analyses on the effectiveness of pretraining. Mahajan et al.~\cite{DMahajanECCV2018} analyze the effect of pretraining when increasing the pretraining dataset size. To significantly increase the dataset size, \cite{DMahajanECCV2018} collect social media images and adopt weakly supervised strategy due to a lack of labels of these images. When dataset scale was extremely enlarged, the classification accuracy was proportionally increased. He et al.~\cite{KHeICCV2019} questioned whether pretraining actually increases the classification accuracy. According to \cite{KHeICCV2019}, a randomly initialized model provides compatible accuracy to the model finetuned from a pretrained model as long as it is trained with extremely large amount of training time. Based on this observation, they conclude that using pretrained models trained on large datasets is not requirement in achieving high accuracy. In this paper we also carry out comprehensive studies on the pretraining for hyperspectral image classification according to recent trends.

\begin{table*}[t]
\caption{{\bf The proposed backbone model architecture.} The table shows the architectural differences between the backbone CNN and the contextual CNN from which the backbone CNN originated. Floating-point operations (FLOPs, multiply \& addition) are calculated assuming that the Indian Pines domain is used. That is, image height (H), image width (W), spectrum dimension, and the number of categories (C) are set as 145, 145, 200, and 9, respectively. In order to have 9 layers in both models, {\bf $\text{k}_1$} and {\bf $\text{k}_2$} are set to 2 and 3, respectively.}

\label{tab:architecture}
\centering
\setlength{\tabcolsep}{14.5pt}
\renewcommand{\arraystretch}{1.2}
\begin{tabular}{c||c|l||c|l}
\specialrule{.15em}{.05em}{.05em}
layer name & output size & contextual CNN~\cite{HLeeTIP2017} & output size & our backbone \\\specialrule{.15em}{.05em}{.05em}
\multirow{4}{*}{Conv 1} & H$\times$W$\times$128 & 1$\times$1, 128 & \multirow{4}{*}{H$\times$W$\times$128} & \multirow{4}{*}{5$\times$5, 128, pad 2} \\
& H$\times$W$\times$128 & 3$\times$3, 128, pad 2$^{\star}$ & & \\
& H$\times$W$\times$128 & 5$\times$5, 128, pad 4$^{\dagger}$ & & \\\cline{2-3}
& H$\times$W$\times$384 & channel-wise concat. & & \\\hline
Conv 2 & H$\times$W$\times$128 & 1$\times$1, 128 & H$\times$W$\times$128 & 1$\times$1, 128 \\\hline
\multirow{3}{*}{Res x} & \multirow{3}{*}{H$\times$W$\times$128} & \multirow{3}{*}{
$\begin{bmatrix}
\text{1$\times$1}, 128\\
\text{1$\times$1}, 128
\end{bmatrix}$
$\times\text{k}_\text{1}$} & \multirow{3}{*}{H$\times$W$\times$128} & \multirow{3}{*}{
$\begin{bmatrix}
\text{1$\times$1}, 128\\
\text{1$\times$1}, 128
\end{bmatrix}$
$\times \text{k}_\text{2}$}\\
&&&&\\
&&&&\\\hline
Classif 1 & H$\times$W$\times$128 & 1$\times$1, 128 & & \\\cline{1-3}
Classif 2 & H$\times$W$\times$128 & 1$\times$1, 128 & & \\\hline
Classif 3 & H$\times$W$\times$C & 1$\times$1, C & H$\times$W$\times$C & 1$\times$1, C \\\specialrule{.15em}{.05em}{.05em}
FLOPs$^{\ddagger}$ & & 43.9$\times \text{10}^\text{9}$~($\text{k}_\text{1}$=2) & & 31.8$\times \text{10}^\text{9}$~($\text{k}_\text{2}$=3) \\
\specialrule{.15em}{.05em}{.05em}
\multicolumn{5}{l}{$\star$ 3$\times$3 max pooling is applied to the output.}\\
\multicolumn{5}{l}{$\dagger$ 5$\times$5 max pooling is applied to the output.}\\
\multicolumn{5}{l}{$\ddagger$ FLOPs is calculated as the equation given in~\cite{HLeeDCS2019}.}\\
\end{tabular}
\end{table*}

\subsection{Hyperspectral Image Classification} 
\label{ssec:hyperspectral_classif}

Recently, many CNN-based approaches have been introduced to tackle hyperspectral image classification. Most of CNN-based approaches have rebuilt the architecture with existing layers or modules used for other typical recognition problems such as image classification (e.g., LeNet~\cite{SMeiTGARS2017}, residual module~\cite{HLeeTIP2017}, multi-scale filter bank~\cite{HLeeTIP2017,ZGongTGARS2019,MZhangTIP2018}, deconvNet exploiting deconvolutional layers~\cite{XMaTGARS2018}, Long Short-Term Memory (LSTM)~\cite{LMouTGARS2017,YXuTGARS2018}, Recurrent Neural Network (RNN)~\cite{XYangTGARS2018}, capsule module~\cite{MPaolettiTGARS2019v2} etc.). Chen al al.~\cite{YChenTGARS2019} introduce a system that automatically designs a structure with the existing layers to provide the highest classification accuracy. Some previous approaches improve accuracy by integrating multiple methods~\cite{XCaoTIP2018} or multiple different features~\cite{GChengTGARS2018,AGuoTGARS2019}.

There are also several literatures that introduce new layers or modules to deal with the inherent limitations of hyperspectral images, especially the overfitting problem that occurs because there are very few hyperspectral images labeled due to the difficulty of annotations (e.g., pyramidal bottleneck module~\cite{MPaolettiTGARS2019}, lightweight unit~\cite{HZhangTGARS2019}, multi-bias layer module~\cite{LFangTGARS2019}, the Generative Adversarial Network (GAN)~\cite{LZhuTGARS2018}). Instead of new layers or modules, new data augmentation strategies based on either GAN~\cite{JFengTGARS2019,HLeeJSTARS2021} or active learning~\cite{JHautTGARS2018} are also used to cope with the overfitting problem. It is also widely used to simply reduce the dimensions of the image spectrum via PCA or spectrum selection to reduce the size of the model, thus alleviating the overfitting problem to some extent~\cite{WZhaoTGARS2016,NHeTGARS2019,LShuTGARS2018}.

Similar to our approach, some works~\cite{HZhangTGARS2019,JYangTGARS2017,LWindrimTGARS2018,CDengTGARS2019} also use pretraining, but do not properly tackle the issues addressed in this paper (no large-scale source domains, different spectral characteristics between different domains). Windrim et al.~\cite{LWindrimTGARS2018} collect multiple source domains and interpolate spectral characteristics of different domains into the shared spectrum. However, due to the large fluctuations in the spectrum, the interpolation cannot restore the missing spectrum information accurately, and this affected accuracy degradation. Other approaches~\cite{HZhangTGARS2019,JYangTGARS2017,CDengTGARS2019} use only one for the source domain, so large-scale source domains are not built. Our cross-domain approach can address such issues which other works miss and comprehensive studies are conducted to verify the effectiveness of our approach to improve the accuracy of hyperspectral image classification.

\section{Methodology}
\label{sec:method}

Our goal is to explore and figure out a way to resolve the issue of not being able to apply a conventional way of pretraining a CNN model for the task of hyperspectral image classification. We first design a state-of-the-art CNN model for this task by employing recently introduced performance-increasing modules (subsection~\ref{ssec:backboneCNN}). Across all the experiments, this model has been set as the backbone model to evaluate the performance with/without the proposed pretrain-finetune approach. Then, we will also describe the newly devised cross-domain pretrain-finetune approach (subsection~\ref{ssec:pretraining_network}). In subsection~\ref{ssec:optimization}, details used for optimizing the cross-domain pretrain-finetune approach are given.

\begin{figure*}[t]
\centering
\includegraphics[width=0.75\linewidth,trim=5mm 5mm 5mm 5mm,clip]{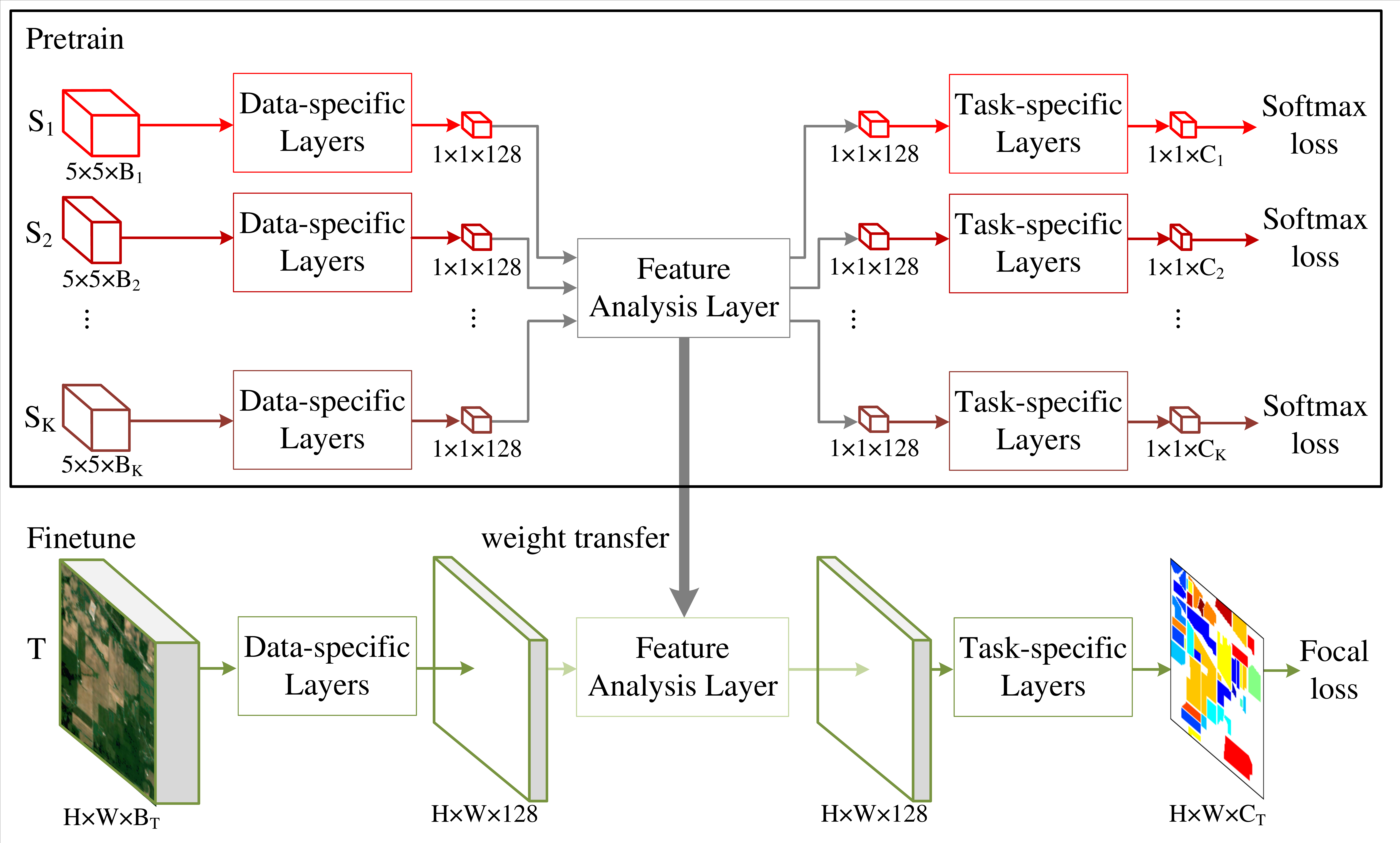}
\caption{{\bf Pretrain-finetune strategy.} A pretrained model is trained with K source domains ($\text{S}_{\text{1}}$, $\cdots$, $\text{S}_{\text{K}}$) and the target model is finetuned from this pretrained model on a target domain T. Feature dimensions are denoted below each blob. For each domain, the number of data channels is denoted by $\text{B}_\text{i}$, and the number of categories classified by the associated task is denoted by $\text{C}_\text{i}$.}
\label{fig:concept}
\end{figure*}

\subsection{Backbone}
\label{ssec:backboneCNN}

The backbone is built based on a contextual CNN architecture which is the deepest model among all existing CNN-based methods~\cite{HLeeIGARSS2016,HLeeTIP2017}. The contextual CNN is a fully convolutional network and contains the sequentially connected multi-scale filter bank, one convolutional layer, two 2-layer residual modules, and three convolutional layers, as shown in Table~\ref{tab:architecture}. Multi-scale filter bank~\cite{CSzegedyCVPR2015} consists of 1$\times$1, 3$\times$3, and 5$\times$5 filters for better analysis of the spatial characteristics. Local response normalization (LRN)~\cite{AKrizhevskyNeurIPS2012} for better generalization during training is applied after the first and second convolutional layers. Dropout layers~\cite{GHintonArXiv2012} were adopted after 7th and 8th layers to enable faster training by preventing complex co-adaptation of each neuron. Each layer consists of 128 convolutional filters.

The backbone was constructed by adopting the contextual CNN with a recently introduced new layer known to improve accuracy while removing ineffective modules, as follows:
\begin{enumerate}[label=(\roman*)]
\item {\bf No multi-scale filter bank:} Instead of a complex multi-scale filter bank, a convolutional layer with only 5$\times$5 filters is used.
\item {\bf More residual modules:} All layers are made up of residual modules, except the first, second, and last layers. Dropout is not necessary because it has the similar functionality as the residual module for easing network training.
\item {\bf Batch normalization (BN):} BN~\cite{SIoffeICML2015} is adopted right after each convolutional layer, so there is no need for another normalization method such as LRN. BN helps the network learn the representation of the entire hyperspectral image when using a small number of training example. By adding BN to the network, bias terms are unnecessary unlike the original CNN architecture.
\item {\bf Focal loss:} We use the focal loss~\cite{TLinICCV2017} to optimize the model for a target task. This loss enables all examples to be effectively considered for each iteration by giving higher weights to harder examples. Since hyperspectral domains usually contain many easy examples, the focal loss was found to be suitable. Softmax loss is used when pretraining on multiple source domains as used in contextual CNN.
\end{enumerate}
Such changes reduce the network size and FLOPs while increasing the accuracy according to our experiments (Table~\ref{tab:new_component}). Table~\ref{tab:architecture} compares the modified backbone with the contextual CNN in terms of architectural differences and FLOPs.

\begin{table*}[t]
\caption{{\bf Specifications for different hyperspectral domains.} We use the domain abbreviations written in parentheses throughout the paper. Reduced bands are acquired by removing the bands which correspond to the water absorption. Note that, datasets acquired using the same type of sensor can have different reduced bands because the deleted bands were chosen according to the task unique to each dataset.}

\label{tab:domains}
\centering
\setlength{\tabcolsep}{7.5pt}
\renewcommand{\arraystretch}{1.2}
\begin{tabular}{l|c|c|c|c|c|c}
\specialrule{.15em}{.05em}{.05em} 
domain & sensor & range & bands & reduced bands & \# data & \# class \\\specialrule{.15em}{.05em}{.05em}
Indian Pines ({\bf I}) & \multirow{3}{*}{AVIRIS} & \multirow{3}{*}{0.4$\sim$2.5$\mu$m} & \multirow{3}{*}{224} & 200 & \multicolumn{1}{r|}{8,504} & 9 \\\cline{1-1}\cline{5-7}
Salinas ({\bf S}) & & & & 204 & \multicolumn{1}{r|}{54,129} & 17 \\\cline{1-1}\cline{5-7}
Kennedy Space Center ({\bf KSC}) & & & & 176 & \multicolumn{1}{r|}{5,211} & 14 \\\hline
Pavia Centre ({\bf PC}) & \multirow{2}{*}{ROSIS} & \multirow{2}{*}{0.43$\sim$0.86$\mu$m} & \multirow{2}{*}{115} & 102 & \multicolumn{1}{r|}{7,456} & 10 \\\cline{1-1}\cline{5-7}
Pavia University ({\bf PU}) & & & & 103 & \multicolumn{1}{r|}{42,776} & 10 \\\hline
Botswana ({\bf B}) & Hyperion & 0.4$\sim$2.5$\mu$m & 242 & 145 & \multicolumn{1}{r|}{3,248} & 15 \\\hline
Houston University ({\bf H}) & Unknown & 0.38$\sim$1.05$\mu$m & 144 & 144 & \multicolumn{1}{r|}{15,029} & 15 \\\specialrule{.15em}{.05em}{.05em} 
\end{tabular}
\end{table*}

\subsection{Building Cross-Domain Pretrained Model}
\label{ssec:pretraining_network}

Our pretrained model is built by using a cross-domain approach. This enables the training of a CNN model which can be universally deployed for various hyperspectral domains. The architecture sequentially consists of multiple inlets (i.e., data analysis layers), shared portion across multiple domains (i.e., mid-level feature analysis layers), and multiple task-specific layers. The data analysis layers that physically handle different datasets cannot be shared universally due to the inconsistent number of channels in each dataset. Task-specific layers that depend on the task properties such as category size cannot be shared as well.

After careful consideration, all residual modules are selected as`` mid-level feature analysis layers'' that are shared by the various spectral domains. The 1st and 2nd layers are treated as ``data-specific layers'' that analyze data-specific spectral characteristics, and the last set of layers are considered as ``task-specific layers''.

In making use of a pretrained model towards a ``target'' dataset, we finetune the middle portion (``mid-level feature analysis layers'') from the pretrained model while other layers are learned from scratch. In the finetune process, we use 10$\times$ learning rate for the data analysis layers (while keeping 1$\times$ for the shared portion). This specific procedure allows the data analysis layers to quickly converge on the target dataset, while slowing down the optimization of the shared portion of the model in order to maintain the ability to extract the general spectral characteristics acquired from the source domain. This was crucial to obtain a significant accuracy gain when compared to not adjusting the learning rate. The overall pretrain-finetune strategy is illustrated in Figure~\ref{fig:concept}.\medskip

\noindent{\bf Reasoning about the role of each layer.} Even though it is challenging to semantically find the role of each layer of CNN model, we can infer these roles. In general, all CNN layers other than the last layer are considered \emph{feature encoder} which encodes input examples in the learned common feature space, and the last layer uses these features to carry out a classification task. If there are multiple domains to be encoded in a common feature space and the characteristics of domains are significantly different, different encoders are required for each domain. In our cross-domain model, we technically create different encoders for different domains, and this was found to be possible by simply sharing the layers across different encoders but uniquely setting the first layers for different domains.\medskip

\noindent{\bf Relation to previous works.} The effectiveness of our design strategies (i.e., multiple task-specific layers, multiple encoders, and transferring ability when properties between source and target domains are different) have been validated in several previous works shown below:

\begin{itemize}
\item {\bf Multiple task-specific layers.} \emph{Multi-task learning}, performing multiple tasks with different layers in a common learned space, is widely used and has presented better accuracy than using multiple single task networks~\cite{HLeeICASSP2018,IKokkinosCVPR2017,HLeeICIP2017,HLeeICASSP2019,SEumICASSP2019}.\medskip

\item {\bf Multiple encoders.} The key component of \emph{Contrastive Language-Image Pretraining (CLIP)}~\cite{ARadfordICML2021}, which has recently been attracting attention due to its remarkable representative ability in the feature space, is to use two encoders to encode the language input and the image input in a common feature space, like our cross-model architecture.\medskip

\item {\bf Transferring ability when properties between source and target domains are different.} \emph{Self-supervised learning}~\cite{KHeCVPR2020,TChenICML2020,JBGrillNeurIPS2020,MCaronNeurIPS2020,XChenCVPR2021,XChenICCV2021}, which has recently been explosively used in representation learning, shows remarkable performance when transferring to target tasks using a pretrained networks on a simple pretext task without relying on class labels. These works show that the pretrain-finetune strategy is effective even when the source task and the target task are completely different.
\end{itemize}

\subsection{Optimization}
\label{ssec:optimization}

We prepare different training strategies depending on the training scenario:
\begin{itemize}
\item {\bf Batch containing the entire set of examples (Target domain):} When training a model with a single target domain, we use a batch containing the entire set of examples for each training iteration to expedite the convergence. Focal loss has a capability to effectively train on such a batch.
\item {\bf Mini-batch built with randomly selected examples (Source domain):} Unlike the target domain scenario, a mini-batch optimization is used. This is due to the fact that the source domain training involves multiple domains concurrently, inevitably providing GPU memory issues when the entire set of examples are used for training.
\item {\bf Two-step cascade training strategy (Source domain):} When using multiple datasets for training, we should consider a potential issue caused by the imbalanced datasets. Since the Salinas dataset ({\bf S}) or the Pavia University ({\bf PU}) has much more data, it requires more iterations than the others. To cope with this issue, we adopt a two-step optimization strategy introduced in~\cite{HLeeICIP2017,HLeeICASSP2019,HLeeICASSP2020}. Under this scheme, the model is initially trained on only the largest dataset (Step I ({\bf S} or {\bf PU})) and then is updated using the whole dataset (Step II). {\bf S} and {\bf PU} are used together as source domains, cross-domain optimization is also used in step I. Both steps use a mini-batch optimization.\medskip
\end{itemize}

\noindent{\bf Initialization.} The layers that are not inherited from the pretrained model are initialized according to the Gaussian with zero-mean and standard deviation of 0.001. This setting was very important in providing high accuracy. When we had adopted recently introduced initialization methods (e.g., Xavier~\cite{XGlorotAISTATS2010}, kaiming\_init~\cite{KHeICCV2015}) or higher standard deviation (e.g., 0.01), the accuracy was significantly degraded by $\sim$30\%.\medskip

\noindent{\bf Data augmentation.} Hyperspectral image classification typically uses a small number of examples for training (e.g., several thousands examples to optimize 1.1M parameters), which can cause overfitting problems. To provide richer set of examples to cope with this issue, the training examples are augmented eight-fold by mirroring each example across the vertical, horizontal, and two diagonal axes~\cite{HLeeTIP2017}.\medskip

\noindent{\bf Learning rate tuning for shared layers.} When learning the shared layers, we multiply $1/N$ (where $N$ is the number of domains involved in the training process) to the learning rate because updating the weights in these layers are affected by all $N$ losses when back-propagation takes place at each iteration. This strategy has proven effective for multi-task learning in~\cite{HLeeTIP2020}.

\section{Experiments}
\label{sec:experiment}

\subsection{Settings}
\label{ssec:eval_setting}

\noindent{\bf Domains.} Seven hyperspectral domains shown in Table~\ref{tab:domains} are used for the experiments. The Indian Pines domain has 17 classes but only 9 classes (including background class) with relatively large numbers of examples are used. For other domains, we use all available classes. Indian Pines domain is used as the target domain while various combinations of the remaining six domains are used as the source domains. Once a pretrained model is built using source domains, finetuning process is carried out on a target domain. 

Our main goal is to explore how pretrain-finetune strategy affects the overall accuracy for the target domain. Therefore, when training on the source domains (i.e., pretraining), all the available examples are used. For the target domain, 200 pixels randomly selected from each category are used for training and all the remaining pixels are used for testing.  All the accuracy values reported hereafter are acquired solely on the target domain.\medskip

\noindent{\bf Learning specifications.} We use the stochastic gradient descend (SGD) approach for training. We set the momentum, gamma, and weight decay as 0.9, 0.1, and 0.0005, respectively.

When {\bf S} or {\bf PU} is not included in the source datasets (i.e., one-step optimization), a pretrained model is trained with a learning rate of 0.01 for 2K iterations. When a two-step cascade optimization is used, we use a learning rate of 0.01 for 1K iterations for the first step, while the second step uses the same settings used in the one-step optimization. For the target domain, the model is trained with a learning rate of 0.001 for 100 iterations.\medskip

\noindent{\bf Evaluation metrics.} We consider three metrics for evaluation: overall accuracy (OA), average accuracy (AA), and kappa static. All three metrics are used when comparing with other previous methods, while only OA is used for the ablation experiments.

\begin{figure}[t]
\centering
\includegraphics[width=\linewidth,trim=30mm 90mm 30mm 90mm, clip]{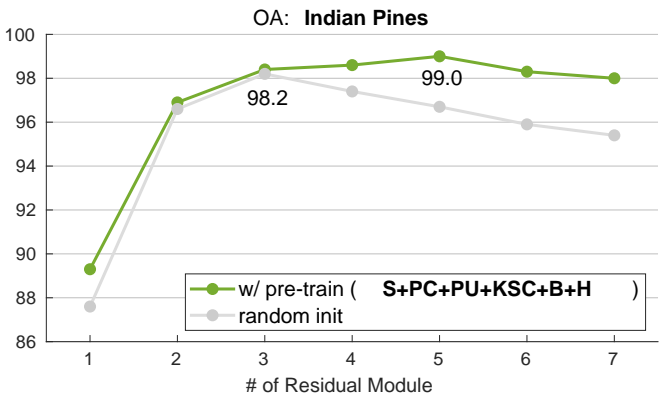}
\caption{{\bf OA on Indian Pines domain with architectures having different depths.} The proposed approach has different architectures depending on the number of the residual modules. The pretrained model is built with six source domains ({\bf S, PC, PU, KSC, B, and H}).}
\label{fig:resmod}
\end{figure}

\begin{figure*}[t]
\centering
  \subfloat[]{
    \includegraphics[width=0.315\textwidth,trim=40mm 86mm 48mm 86mm,clip]{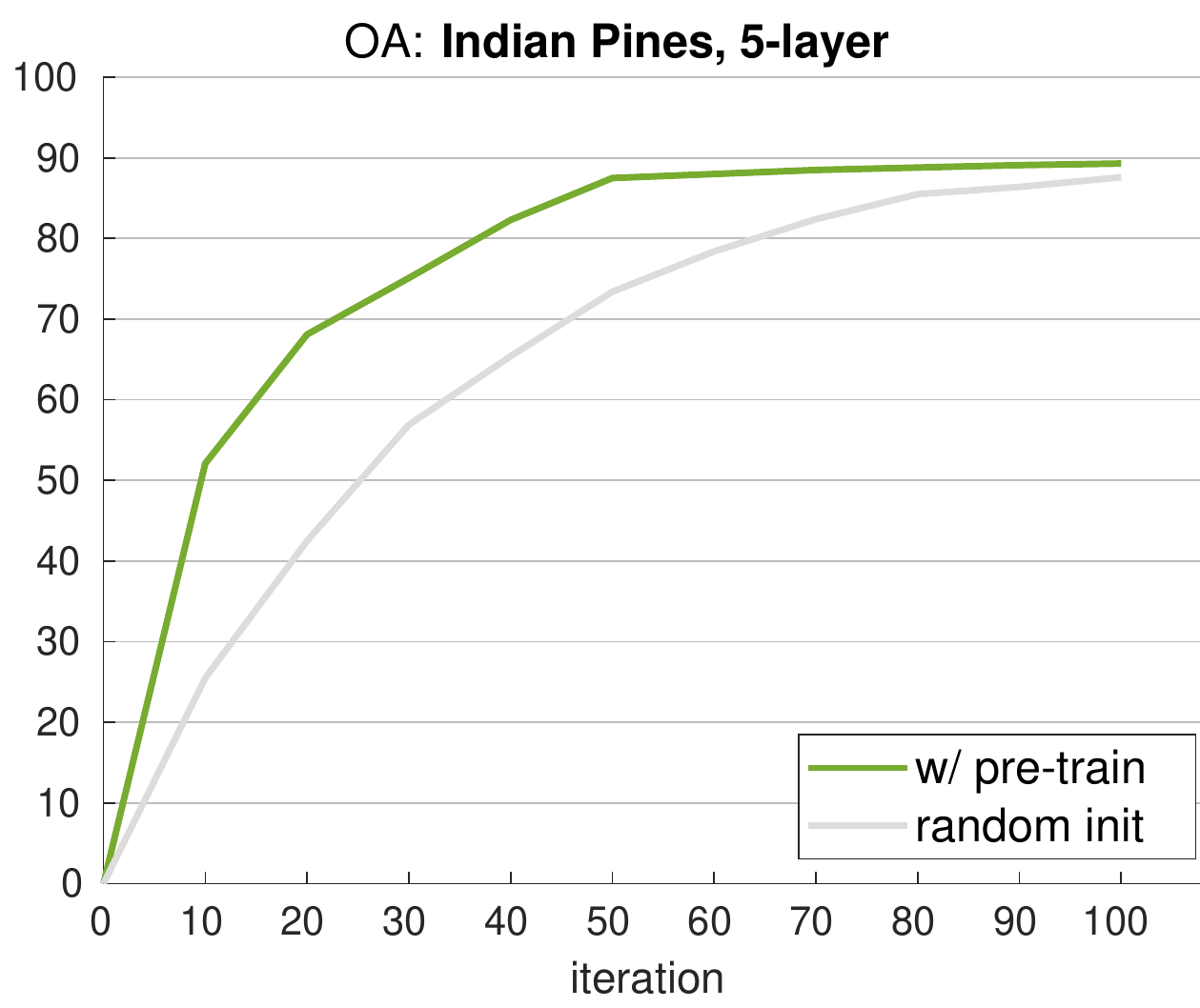}
  }\hfill
  \subfloat[]{
    \includegraphics[width=0.315\textwidth,trim=40mm 86mm 48mm 86mm,clip]{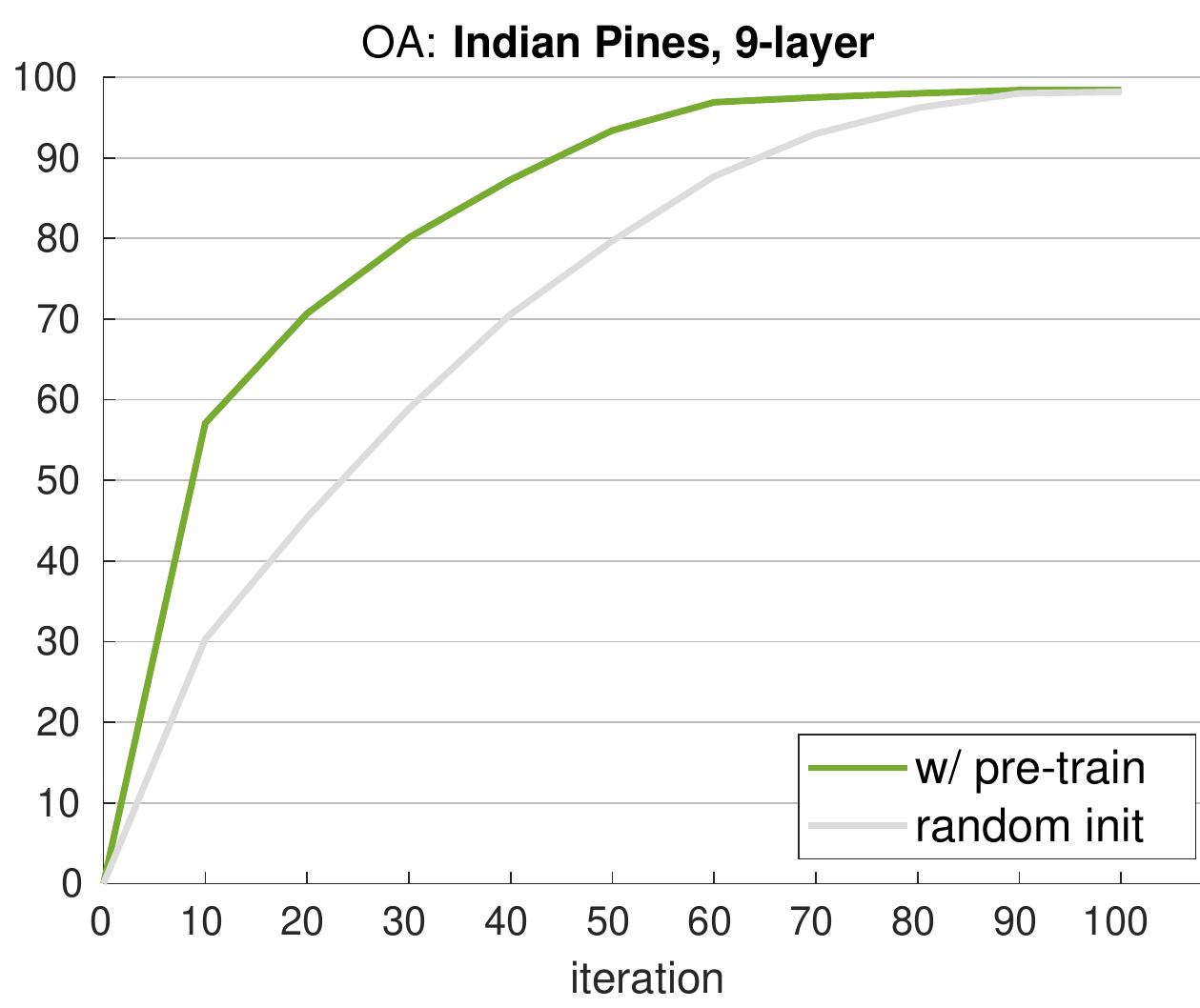}
  }\hfill
  \subfloat[]{
    \includegraphics[width=0.315\textwidth,trim=40mm 86mm 48mm 86mm,clip]{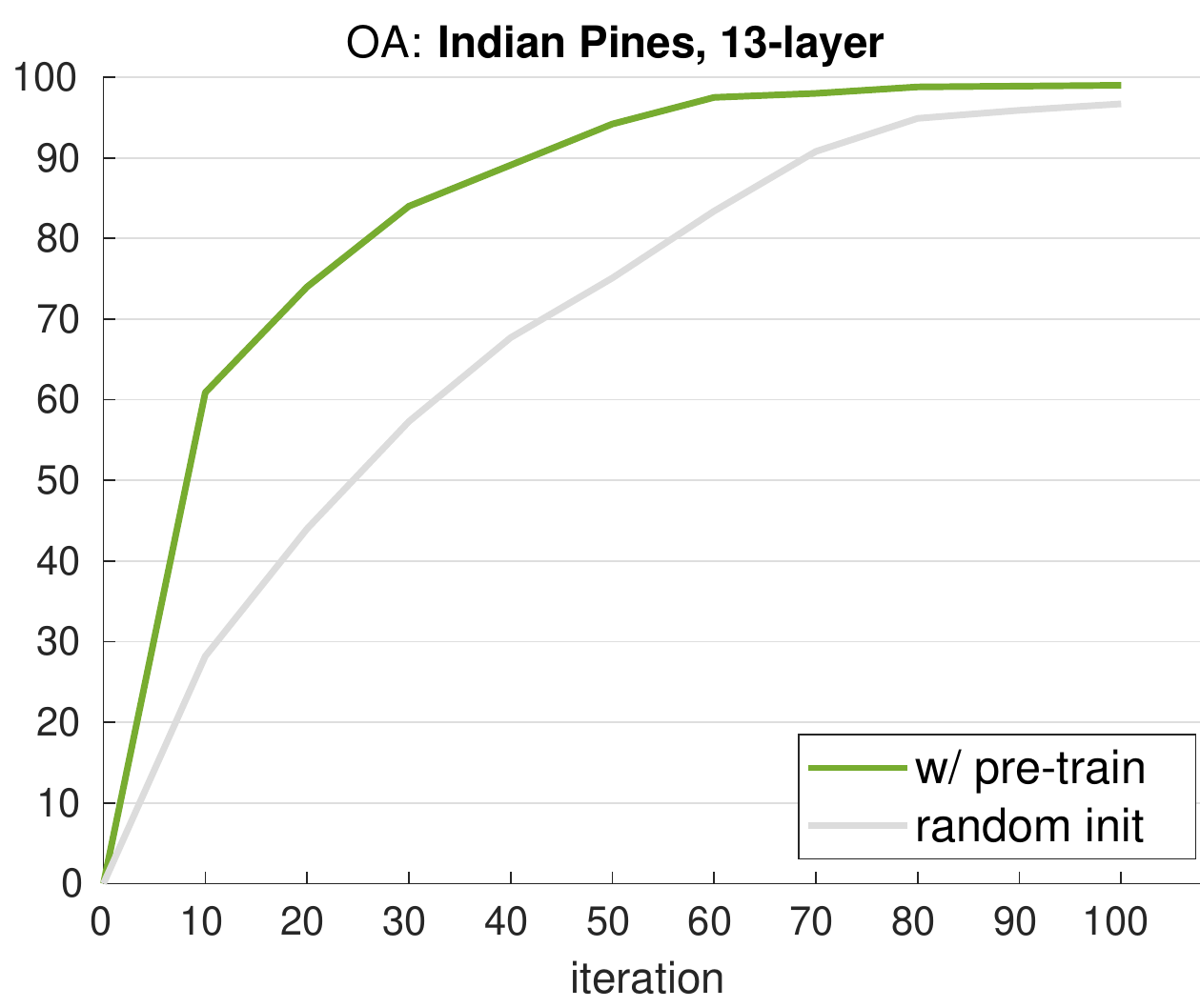}
  }
\caption{{\bf Learning curves on OA with three models (5-layer, 9-layer, 13-layer).} Pretrained models are trained on six source domains and target model is evaluated on the Indian Pines domain.}
\label{fig:map_evolution}
\end{figure*}

\subsection{Analysis of Pretrained Model}
\label{ssec:analysis}

In this subsection, we answer each question claimed in the introduction section and provide relevant experiments to support our answer.\medskip

\noindent{\bf Is pretraining necessary? Yes.} There are three benefits to using a pretrained model as below:
\begin{enumerate}[label=(\roman*)]
    \item Models with a pretrained model consistently provide higher accuracy than their random initialization counterparts, regardless of model architecture.
    \item Model using a pretrained model can be built deeper with more residual modules up to 13 layers without compromising the accuracy.
    \item Learning converges faster, which reduces the training time.
\end{enumerate}

Figure~\ref{fig:resmod} compares the accuracy of models using pretraining and models trained from scratch. Using pretraining consistently provides 0.2\%$\sim$2.6\% higher accuracy than all of the randomly initialized models. 

It is also observed that the 9-layer model provides the highest accuracy when trained from scratch, which is also shown in~\cite{HLeeTIP2017}. However, with a pretrained model, the 13-layer model provides the best accuracy. This ``going deeper'' (i.e., network deepening) was made possible since the overfitting issue could be addressed by finetuning on a pretrained model that has the same effect as increasing the size of the training dataset.

Figure~\ref{fig:map_evolution} shows learning curves for three models (5-, 9-, 13-layer). Models that use pretraining are generally shown to converge faster than their counterparts. Furthermore, a model trained from scratch takes more time to converge as the model goes deeper. This may be due to the expansion of overfitting issues as the model size increases.

Interestingly, our observation is very different from that of~\cite{KHeICCV2019}. For the object detection described in~\cite{KHeICCV2019}, when using a pretrained model, fast convergence was observed, but this model does not outperform the random initialization counterpart regarding to the accuracy. The major claim made in~\cite{KHeICCV2019} is that given sufficient iterations, the model trained from scratch can achieve comparable accuracy as the model using pretraining. It is noteworthy to mention that the same trend was said to have happened when only small number of training examples (e.g., subset of COCO dataset) were available.

However, unlike the claim made in~\cite{KHeICCV2019}, the performance comparison shows that a pretrained model indeed outperforms the one trained from scratch leading to an opposite conclusion. We consider the hyperspectral image classification as an extreme case where very few examples are available for training, thus showing a great demand to expand the training dataset.\medskip

\begin{figure}[t]
\centering
  \includegraphics[width=0.290\textwidth,trim=6mm 30mm 120mm 25mm,clip]{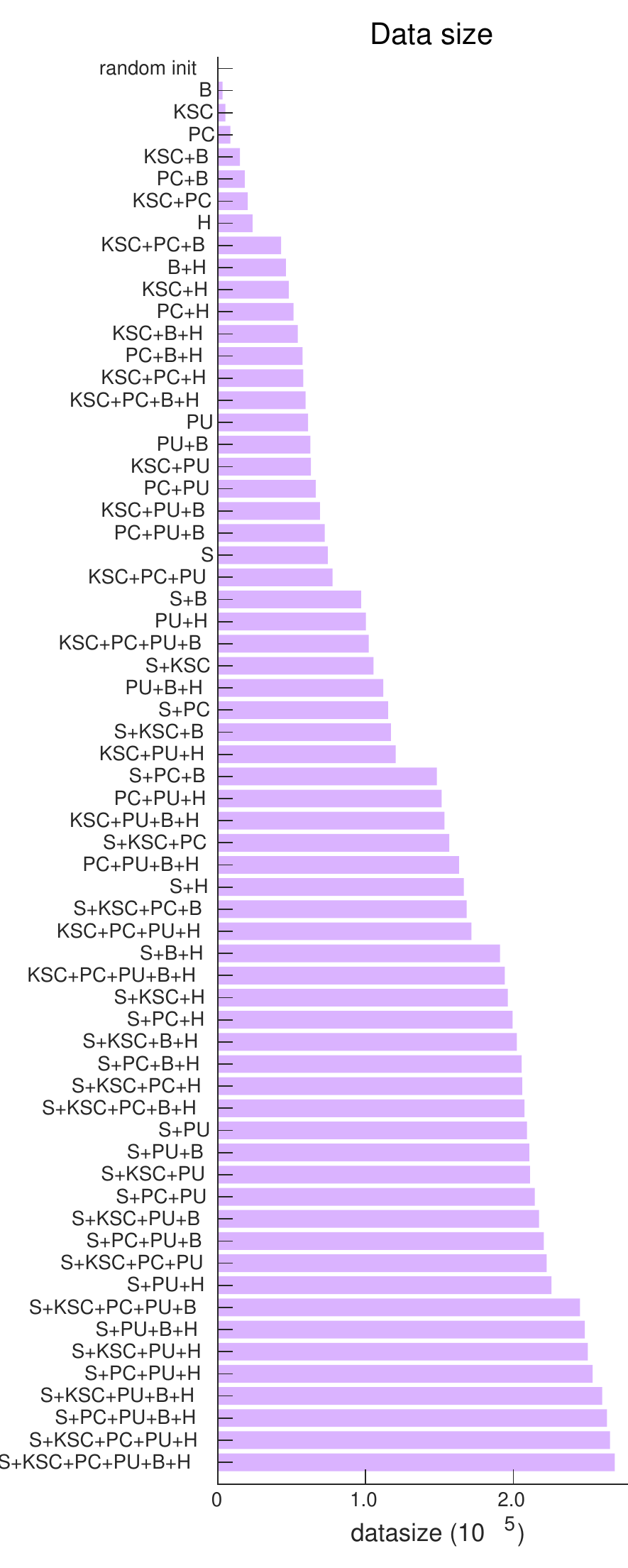}
  \includegraphics[width=0.190\textwidth,trim=125mm 30mm 32mm 25mm,clip]{datasize_vs_OA.pdf}
\caption{{\bf Data size and OA for all combinations of 6 source datasets.} Datasets are sorted according to their sizes and listed on the x axis. Two graphs share the x axis..}
\label{fig:map_vs_datasize}
\end{figure}

\noindent{\bf Does a larger source dataset improve accuracy? Yes.} We question how much influence the dataset size of a source domain has over the classification accuracy. To answer this question, we generate 63 source domains by combining up to 6 source domains and use them to train separate models. Figure~\ref{fig:map_vs_datasize} shows the sizes of 63 source domains (along with one constructed with random initialization) and OAs obtained by finetuning each pretrained model on the target domain. In the figure, we observed that using a larger source domain generally improves the classification accuracy. In all the source cases, accuracy was better than the case where the model was randomly initialized.\medskip

\begin{table*}[t]
\caption{{\bf Comparison w.r.t. source data sensors.} The accuracy of the models using single source domain is shown on the left side. Several combinations of source domains which includes or excludes {\bf S} or {\bf KSC} are constructed and tested on the right side. Note that {\bf S} and {\bf KSC} are taken from the same sensor as {\bf I}.}
\label{tab:different_sensor}

\centering
\setlength{\tabcolsep}{12.0pt}
\renewcommand{\arraystretch}{1.2}
\begin{tabular}{c|cccccc|cccc}
\specialrule{.15em}{.05em}{.05em}
& {\bf S} & {\bf KSC} & {\bf PC} & {\bf PU} & {\bf B} & {\bf H} & {\bf S+} & {\bf KSC+} & {\bf S+} $\cup$ {\bf KSC+} & /({\bf S+} $\cup$ {\bf KSC+}) \\\specialrule{.15em}{.05em}{.05em}
OA & 96.9 & 97.5 & 96.9 & 97.0 & 96.6 & 96.7 & 98.2 & 97.9 & 97.9 & 97.1 \\\hline
data size & 52.8k & 5.1k & 7.3k & 41.8k & 3.2k & 14.7k & $\cdot$ & $\cdot$ & $\cdot$ & $\cdot$ \\\specialrule{.15em}{.05em}{.05em}
\multicolumn{10}{l}{{\bf +}: all sets including the corresponding source dataset (shown before the {\bf +} mark)}\\
\multicolumn{10}{l}{$\cup$: union of two sets}\\
\multicolumn{10}{l}{/( ): a complement set of ( )}
\end{tabular}
\end{table*}

\begin{table*}[t]
\caption{{\bf Performance comparison of using single source and multiple sources.} There are five cases where a single domain and a combination of multiple domains have similar data size in figure~\ref{fig:map_vs_datasize}. Domains are sorted according to the data size so the larger size is placed on the right. In each case, the best accuracy is written in bold.}

\label{tab:single_vs_multiple}
\centering
\setlength{\tabcolsep}{5.5pt}
\renewcommand{\arraystretch}{1.2}
\begin{tabular}{c|cc}
\specialrule{.15em}{.05em}{.05em}
& {\bf PC} & {\bf KSC+B} \\\specialrule{.15em}{.05em}{.05em}
OA & 96.9 & {\bf 97.3} \\\hline
data size & 7.3k & 8.3k
\\\specialrule{.15em}{.05em}{.05em}
\end{tabular}
~~
\begin{tabular}{ccc}
\specialrule{.15em}{.05em}{.05em}
{\bf KSC+PC} & {\bf H} & {\bf KSC+PC+B}  \\\specialrule{.15em}{.05em}{.05em}
96.9 & 96.7 & {\bf 97.9} \\\hline
12.4k & 14.7k & 15.5k
\\\specialrule{.15em}{.05em}{.05em}
\end{tabular}
~~
\begin{tabular}{ccc}
\specialrule{.15em}{.05em}{.05em}
{\bf KSC+PC+B+H} & {\bf PU} & {\bf PU+B} \\\specialrule{.15em}{.05em}{.05em}
{\bf 97.2} & 97.0 & 97.0 \\\hline
30.2k & 41.8k & 44.9k
\\\specialrule{.15em}{.05em}{.05em}
\end{tabular}
~~
\begin{tabular}{ccc}
\specialrule{.15em}{.05em}{.05em}
{\bf PC+PU+B} & {\bf S} & {\bf KSC+PC+PU} \\\specialrule{.15em}{.05em}{.05em}
{\bf 97.3} & 96.9 & 97.0 \\\hline
52.2k & 52.9k & 54.1k
\\\specialrule{.15em}{.05em}{.05em}
\end{tabular}
\end{table*}

\noindent{\bf Does source domain need to come from same sensor as target domain? No.} Pretraining was effective regardless of the sensor type by which the source domain was obtained.

We verify the effect of pretraining a model on a source domain acquired by a sensor different from the target data. In general, using a pretrained model to finetune is effective only if the source data which is used for pretraining and the target data share the same sensor.

As shown in Table~\ref{tab:different_sensor}, using source domains that share the same sensor as the target domain (i.e., {\bf S} and {\bf KSC}) did not provide the benefit of improving the accuracy compared to the cases where the source and the target sensors were different (i.e., {\bf PC}, {\bf PU}, {\bf B}, and {\bf H}). Instead, accuracy seems to be more affected by the dataset size of the domain. In addition, we have tested to validate how the inclusion/exclusion of source data acquired by specific sensors (i.e., matching the target sensor with the source) affect the accuracy when considering combinations of domains. When comparing groups consisting of source domains containing either {\bf S} or {\bf KSC} with a group excluding both domains, the latter group (/({\bf S+} $\cup$ {\bf KSC+})) offers the worst accuracy among all groups, but the difference is marginal (0.8\%).\medskip

\noindent{\bf Does introducing more variety in the source domains for pretraining always increase the accuracy? No.} The source training domain can be either a single domain or a combination of multiple domains. In Figure~\ref{fig:map_vs_datasize}, there are four groups where a single domain and combinations of multiple domains have similar data size. The selected groups are analyzed in detail in Table~\ref{tab:single_vs_multiple}. In all four groups, introducing multiple sources did not bring forth any drastic performance increase. This demonstrates that the accuracy is not affected by the number of different source domains involved in the pretraining.

\subsection{Ablation Study}

We carried out several ablation experiments to validate each component (e.g., cross-domain approach, new network components, etc.) used to analyze the need for a pre-trained model.\medskip

\noindent{\bf Train on source and test on source: effectiveness of cross-domain pretraining.} The cross-domain approach used in the pretraining process has the ability to train a model simultaneously in multiple hyperspectral domains with different spectral characteristics. In this section, we evaluate the effectiveness of the proposed cross-domain pretraining approach (leaving out the finetuning process) when trained and tested on the source domain (source-to-source). Note that this setting is different from the case shown in the main article where pretraining is carried out on the source domain and finetuned on the target domain (source-to-target).

We evaluate whether building a single cross-domain pretrained model which handles multiple domains altogether is better than the models separately trained for specific domains. We use seven domains (i.e., {\bf I}, {\bf S}, {\bf KSC}, {\bf PC}, {\bf PU}, {\bf B}, and {\bf H}). Since {\bf PC} is considered, we use two-stage cascaded optimization approach according to the criteria provided in the main article. For each domain, we use 200 examples randomly chosen from each positive category for training and the remaining for testing. OAs reported in this experiment are the average value over 5 runs. As shown in Table~\ref{tab:accuracy}, our cross-domain approach consistently outperforms the individual models on the Indian Pines, Salinas, KSC, Pavia Center, Pavia University, Botswana, and Houston by 1.1\%, 1.1\%, 1.3\%, 1.7\%, 2.3\%, 1.6\%, and 1.4\%, respectively.

We also compare two cases (source-to-source vs. source-to-target) which are both tested on Indian Pines dataset. For both cases, the dataset is divided into train and test splits using the same regime. We observe that the source-to-source accuracy was higher than that of the source-to-target (99.3 vs. 98.9). This strategy of co-training multiple domains together within source-to-source setting, can be another effective solution to address the overfitting issue caused by insufficient amount of data. However, it suffers from the fact that large memory and multi-domain dataset should be available for training.\medskip

\begin{table}
\caption{{\bf OA on seven domains.} For each domain, model is trained separately from the other domains (individual) or simultaneously with other domains (cross-domain).}
\label{tab:accuracy}

\centering
\setlength{\tabcolsep}{17.2pt}
\renewcommand{\arraystretch}{1.2}
\begin{tabular}{l|c|c|c}
\specialrule{.15em}{.05em}{.05em} 
dataset & individual & cross-domain & gain \\\specialrule{.15em}{.05em}{.05em}
{\bf I} & 98.2 & {\bf 99.3} & +1.1 \\
{\bf S} & 97.6 & {\bf 98.7} & +1.1 \\
{\bf KSC} & 96.6 & {\bf 97.9} & +1.3 \\
{\bf PC} & 96.9 & {\bf 98.6} & +1.7 \\
{\bf PU} & 95.1 & {\bf 97.4} & +2.3 \\
{\bf B} & 97.5 & {\bf 99.1} & +1.6 \\
{\bf H} & 96.9 & {\bf 98.3} & +1.4 \\\specialrule{.15em}{.05em}{.05em} 
\end{tabular}
\end{table}

\noindent{\bf Effectiveness of backbone architecture.} The backbone used in our experiments incorporates various new components that has been proved to be effective in increasing the performance. Table~\ref{tab:new_component} shows the list of components along with the resulting accuracy when each of those were added to the model.

\begin{table*}[t]
\caption{{\bf Newly adopted components of the proposed model.} The proposed model was implemented by adding the recently introduced components to the contextual CNN~\cite{HLeeTIP2017}.}
\label{tab:new_component}

\centering
\setlength{\tabcolsep}{6.5pt}
\renewcommand{\arraystretch}{1.2}
\begin{tabular}{r|cc|ccccccc|c}
\specialrule{.15em}{.05em}{.05em}
& \multicolumn{2}{c|}{contextual~\cite{HLeeTIP2017}} & & & & & & & & the proposed \\\specialrule{.15em}{.05em}{.05em}
no multi-scale filters? & & & \checkmark & & \checkmark & \checkmark & \checkmark & \checkmark & \checkmark & \checkmark \\
more residual modules? & & & & \checkmark & \checkmark & \checkmark & \checkmark & \checkmark & \checkmark & \checkmark \\
BN? & & & & & & \checkmark & \checkmark & \checkmark & \checkmark & \checkmark \\
focal loss? & & \checkmark & & & & & \checkmark & \checkmark & \checkmark & \checkmark \\
w/ pre-train? & & & & & & & & \checkmark & \checkmark & \checkmark \\
initial layer 10$\times lr$ & & & & & & & & & \checkmark & \checkmark \\
shared layer $lr/N$? & & & & & & & & & & \checkmark  \\\specialrule{.15em}{.05em}{.05em}
OA & 93.6 & 93.7 & 93.9 & 96.7 & 96.9 & 97.3 & 98.0 & 98.3 & 98.4 & {\bf 99.0} \\\specialrule{.15em}{.05em}{.05em}
\end{tabular}
\end{table*}

\begin{table}[t]
\caption{{\bf Ablation experiments for focal loss (FL)}. 9-layer model initialized from scratch is used. (a) for each $\gamma$ value, the optimal value of $\alpha$ is used. (b) FL uses the entire set of training sample as a single batch, while random and OHEM use mini-batches with specific batch sizes. For random/OHEM, softmax loss is used. Time indicates the relative wall-clock training time w.r.t. the case of FL, and was measured until the highest accuracy was achieved. Since each iteration time is different for the three methods (random, OHEM, and FL), the wall clock training time is used instead of the total iteration.}

\centering
\setlength{\tabcolsep}{8.0pt}
\renewcommand{\arraystretch}{1.2}
\subfloat[Varying $\gamma$ and $\alpha$]{
\begin{tabular}{cc|c}
\specialrule{.15em}{.05em}{.05em}
$\gamma$ & $\alpha$ & OA \\\specialrule{.15em}{.05em}{.05em}
0 & .75 & 98.0 \\
0.1 & .75 & 98.0 \\
0.2 & .75 & 98.0 \\
0.5 & .50 & 98.2 \\
1.0 & .25 & 98.3 \\
2.0 & .25 & 98.9 \\
5.0 & .25 & {\bf 99.0} \\\specialrule{.15em}{.05em}{.05em}
\end{tabular}
\label{tab:fl_param}
}\hfill
\subfloat[Batch Method]{
\begin{tabular}{c|c|c|c}
\specialrule{.15em}{.05em}{.05em}
method & batch size & OA & time \\\specialrule{.15em}{.05em}{.05em}
\multirow{3}{*}{Random} & 128 & 98.0 & $\times$2.4 \\
& 256 & 97.6 & $\times$2.1 \\
& 512 & 97.6 & $\times$1.6 \\\hline
\multirow{3}{*}{OHEM} & 128 & 98.2 & $\times$3.8\\
& 256 & 98.2 & $\times$3.7\\
& 512 & 97.9 & $\times$3.4\\\hline
{\bf FL} & 1800 & {\bf 99.0} & \\\specialrule{.15em}{.05em}{.05em}
\end{tabular}
\label{tab:batch_method}
}
\end{table}

There are three architectural modifications: no multi-scale filters, more residual modules, and BN. Using a single 5$\times$5 convolutional layer instead of a complicated multi-scale filters can improve performance slightly (0.3\%) and it simplifies the architecture with fewer parameters. The other two modifications serve to increase the accuracy with a large gain of 3.1\% and 3.3\%, respectively. Easing model optimization by incorporating more residual modules seems to be very effective in improving performance. From the observation that BN also provides improved performance, better regularization by normalizing output of each layer over the examples in each batch was effective.

Focal loss was not very effective in improving accuracy. When using focal loss for training \cite{HLeeTIP2017} and our backbone, the accuracy change was 0.1\% and 0.8\%, respectively. The benefit of using focal loss is faster convergence than using other loss types. This will be analyzed further in the next subsection.

The pretrained model was effective in improving performance only when using a large learning rate for the initial layers. Without such adjustment, it is observed that accuracy was rather decreased. Adjusting the learning rate for the shared layers (i.e., shared layer {\it lr}/{\it N}) provides a further improvement of 0.1\%.\medskip

\noindent{\bf Analysis of the use of a focal loss.} As described in~\cite{TLinICCV2017}, two hyper-parameters of focal loss, $\gamma$ and $\alpha$ control the strength of the modulation term. $\alpha$ is used for addressing the positive/negative imbalance problem, while $\gamma$ gives more weight to difficult training examples. OA on the target domain (i.e., Indian pines) for various $\gamma$ and $\alpha$ are shown in Table~\ref{tab:fl_param}. For this comparison, we use 9-layer model initialized from scratch. The highest accuracy was achieved when taking a high $\gamma$ (5.0) and a low $\alpha$ (0.25). This co-aligns with the fact that Indian Pines domain has many easy examples and the low ratio between foreground and background examples. The selected numbers for the two parameters are used throughout all the experiments carried out in the main article.

In table~\ref{tab:batch_method}, we compare the case where we train the entire set of examples at once (using focal loss) to the case where mini-batch-based optimization approaches are used (e.g., random selection, online hard example mining (OHEM)~\cite{AShrivastavaCVPR2016}) in terms of accuracy and training time. While the focal loss is used to process the batch consisting of the entire examples, mini-batch-based optimization uses a softmax loss. Using a focal loss provides higher accuracy by at least 0.8 and was effective in reducing training time by at least $\times$1.6 compared to other mini-batch-based methods. Among mini-batch-based optimization methods, OHEM generally provides higher accuracy but requires longer training time than random selection. Both random selection and OHEM do not greatly depend on batch size when considering OA.\medskip

\begin{figure}[t]
\centering
\includegraphics[width=\linewidth,trim=10mm 95mm 15mm 95mm,clip]{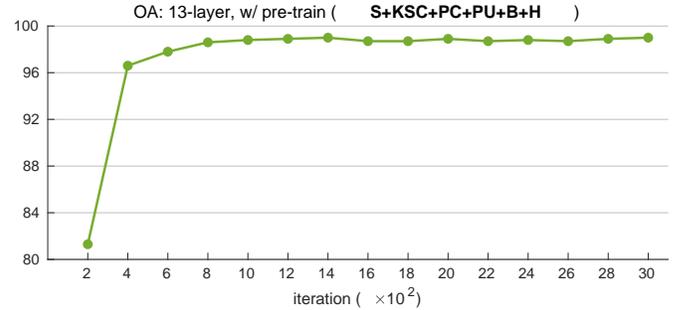}
\caption{{\bf Evolution w.r.t. pretraining schedule.} We have tested various pretrained models using different schedule.}
\label{fig:ap_pretrain_iteration}
\end{figure}

\noindent{\bf Appropriate training schedule for a pretrained model.} To find the proper training schedule, we compare various pretrained models with different training iterations in terms of OA. Figure~\ref{fig:ap_pretrain_iteration} shows the OA with various pretraining iterations. The highest accuracy was achieved at 2k iterations and did not show any additional improvement thereafter. Accordingly, we have selected 2k as the number of iterations to acquire our pretrained model.

\subsection{Comparison with Baselines}
\label{ssec:state-of-the-art}

\begin{table}[t]
\caption{{\bf Comparisons with CNN-based baselines.} For all the baseline models except SSRN and HybridSN, we use DeepHyperX implementations~\cite{NAudebertGithub2019}. For SSRN and HybridSN, we have used the original code provided by the authors.}
\label{tab:state_of_the_art}

\centering
\setlength{\tabcolsep}{8.0pt}
\renewcommand{\arraystretch}{1.2}
\begin{tabular}{l|cc|cccc}
\specialrule{.15em}{.05em}{.05em}
method & layer & \# params. & OA & AA & $k$ \\\specialrule{.15em}{.05em}{.05em}
Semi-super 1D~\cite{ABoulchGRETSI2017} & 10 & \multicolumn{1}{r|}{6.7K} & 89.1 & 87.6 & 84.9 \\
1D~\cite{WHuJS2015} & 2 & \multicolumn{1}{r|}{69.9K} & 90.1 & 89.6 & 86.9 \\
3D~\cite{YChenTGARS2016} & 4 & \multicolumn{1}{r|}{1.0M} & 90.3 & 90.6 & 87.4 \\
2D~\cite{VSharmaTR2016} & 5 & \multicolumn{1}{r|}{2.5M} & 92.8 & 90.3 & 85.9 \\
Contextual~\cite{HLeeTIP2017} & 9 & \multicolumn{1}{r|}{1.0M} & 93.6 & 95.8 & 94.7 \\
3D-2D~\cite{YLuoICALIP2018} & 4 & \multicolumn{1}{r|}{99.0M} & 93.8 & 92.8 & 89.8 \\
3D-1D~\cite{AHamidaTGARS2018} & 10 & \multicolumn{1}{r|}{38.8K} & 94.3 & 93.8 & 89.7 \\
Semi-super 2D~\cite{BLiuRSL2017} & 6 & \multicolumn{1}{r|}{4.6M} & 94.6 & 96.2 & 92.4 \\
SSRN~\cite{ZZhongTGARS2018} & 11 & \multicolumn{1}{r|}{717.8K} & 95.8 & 97.1 & 93.2 \\
2-stream 3D~\cite{YLiRS2017} & 3 & \multicolumn{1}{r|}{78.7K} & 96.5 & 94.9 & 89.7 \\
Multi-scale 3D~\cite{MHeICIP2017} & 10 & \multicolumn{1}{r|}{169.7K} & 96.6 & 93.8 & 90.4 \\
HybridSN~\cite{SKumarGRSL2020} & 7 & \multicolumn{1}{r|}{4.8M} & 96.8 & 97.7 & 95.3 \\\hline
Our backbone & 13 & \multicolumn{1}{r|}{1.1M} & 98.0 & 98.9 & 97.7 \\
~~~~w/ pretraining & 13 & \multicolumn{1}{r|}{1.1M} & {\bf 98.9} & {\bf 99.3} & {\bf 98.1} \\\specialrule{.15em}{.05em}{.05em}
\end{tabular}
\end{table}

We have compared our model with the CNN-based baselines that can be trained in an end-to-end fashion. To make a fair comparison, bells and whistles that cannot be used to carry out the end-to-end training have been omitted in all the models. Table~\ref{tab:state_of_the_art} compares our model with the selected baselines using the evaluation metrics of OA, AA, and $k$. Our backbone model even without finetuning is better than the baselines by at least 1.2\% for OA, 1.2\% for AA, and 2.4\% for $k$. When adopting the pretrain-finetune strategy to utilize representative capability of a pre-trained model on the six source domains, the accuracy was further enhanced by 0.9\% for OA, 0.4\% for AA, and 0.4\% for $k$. We also confirmed that our model has the deepest architecture without compromising the accuracy.

\subsection{Proof of Concept}

\noindent{\bf Representation capability.} We carried out an additional experiment to evaluate the representation capability of pretrained network without finetuning. For this evaluation, the layers transferred from a pretrained network is fixed with the weights trained on source domains (i.e., {\bf S}+{\bf KSC}+{\bf PC}+{\bf PU}+{\bf B}+{\bf H}), but the remaining layers (data-specific layer and task-specific layer) are trained on the novel classes defined in the target task. As shown in the Table~\ref{tab:representation_capability}, accuracy achieved without finetuning was much better than ``random init.'' by a significant margin and comparable to other models without/with the pretrain-finetune strategy in which all layers are updated during training. This comparison supports the representation capability of the pretrained model.\medskip

\begin{table}[t]
\caption{{\bf Effectiveness of pretrain/finetune.} ``random init.'' is the model in which the layers except data-specific layer and task-specific layer are randomly initialized and not updated during training.}
\label{tab:representation_capability}

\centering
\setlength{\tabcolsep}{6.8pt}
\renewcommand{\arraystretch}{1.2}
\begin{tabular}{c|c|ccc}
\specialrule{.15em}{.05em}{.05em}
method & random init. & w/o pretrain & w/o finetune & w/ finetune \\\specialrule{.15em}{.05em}{.05em}
OA & 15.0 & 98.0 & 87.3 & 98.9 \\\specialrule{.15em}{.05em}{.05em}
\end{tabular}
\end{table}

\noindent{\bf Other source domain?} We carried out further experiments considering other datasets (Salinas and the Botswana dataset) as target domains. Two selected datasets have distinct inherent properties compared to other datasets as the Salinas dataset contains the largest number of labeled examples and the Botswana dataset was taken from the satellite unlike other datasets taken from airborne. When acquiring the pretrained model (source domain), we use all the datasets except the one used as the target domain. As shown in Table~\ref{tab:other_sources}, using the pretrain-finetune strategy consistently provides better accuracy than its counterpart of training-from-scratch, regardless of the source domain. Margins achieved by the pretrain-finetune strategy were 0.5\% and 0.9\% for Salinas and Botswana dataset, respectively. When the source domain is a large-scale (e.g., {\bf S}), the impact of using the pretrained model seems to be reduced. On the other hand, the fact that the source and target domains are from different sources between satellites or airborne (e.g., {\bf B}) does not seem to significantly affect the effectiveness of the pretrain-finetune strategy.

\begin{table}[t]
\caption{{\bf Accuracy of the pretrain-finetune strategy on Salinas ({\bf S}) and Botswana ({\bf B}).} When a pretrained model is used, all domains except the target domain are used as source domains.}
\label{tab:other_sources}

\centering
\setlength{\tabcolsep}{16.5pt}
\renewcommand{\arraystretch}{1.2}
\begin{tabular}{c|c|c|c}
\specialrule{.15em}{.05em}{.05em}
domain & w/o pretrain & w/ pretrain & gain \\\specialrule{.15em}{.05em}{.05em}
{\bf S}& 97.6 & {\bf 98.1} & +0.5 \\
{\bf B} & 97.3 & {\bf 98.2} & +0.9 \\
\specialrule{.15em}{.05em}{.05em}
\end{tabular}
\end{table}

\section{Conclusion}
\label{sec:conclusion}

The conventional pretrain-finetune strategy cannot be directly deployed in hyperspectral domains due to three major issues: 1) inconsistent spectral characteristics among the domains, 2) inconsistent number of data channels among the domains, and 3) absence of large-scale domains. We have devised a cross-domain model that can be universally deployed for various hyperspectral domains. The proposed architecture has multiple inlets and outlets that handle different domains and classification tasks, with the middle portion shared by all domains. The middle portion is expected to extract the universal spectral properties from all the domains involved in training and to transfer them to the target task in the finetune process.

We carried out a comprehensive study using a variety of ablation experiments to confirm the effectiveness of this pretrain-finetune strategy. Our experiments demonstrate the benefits of using pretraining in three aspects: i) increasing accuracy, ii) building deeper models without sacrificing the performance, and iii) providing faster training convergence which results in reducing training time. From the experiments, we have also observed that the accuracy of the target task does not depend on sensor types or the number of source domains, but rather on the data size of the combined source domain.

\section*{Acknowledgment}

We would like to thank Prof. Wonkook Kim at Pusan National University for his help with the experiments.


\ifCLASSOPTIONcaptionsoff
  \newpage
\fi



\bibliographystyle{IEEEtran}
\bibliography{references.bib}
\end{document}